\documentclass[10pt]{article}

\usepackage[preprint]{tmlr}
\usepackage{amsmath,amssymb,amsthm,array,booktabs,graphicx,hyperref,microtype}
\usepackage{caption}
\usepackage{float}
\usepackage[section]{placeins}
\usepackage{needspace}
\usepackage{xcolor}
\newcommand{\coderepo}{\url{https://github.com/aqibrahimbt/conformal_adaptive_operators}}
\graphicspath{{figures/}}

\hypersetup{colorlinks=true,linkcolor=blue,citecolor=blue,urlcolor=blue}
\title{Conformally Calibrated Adaptive Computation\\
in Neural Operators}
\author{\name Abdul Qadir Ibrahim \email abdul.qadir.ibrahim@desy.de\\
      \addr Deutsches Elektronen-Synchrotron DESY, Hamburg, Germany
      \AND
      \name Martin Burger \email martin.burger@desy.de\\
      \addr Deutsches Elektronen-Synchrotron DESY, Hamburg, Germany
      \AND
      \name Abel Maviano Jonathan \email abelmaviano.jonathan@graduate.univaq.it\\
      \addr Department of Information Engineering, Computer Science and Mathematics\\
      University of L'Aquila, L'Aquila, Italy}
\hypersetup{pdfauthor={Abdul Qadir Ibrahim, Martin Burger, Abel Maviano Jonathan},pdftitle={Conformally Calibrated Adaptive Computation in Neural Operators},pdfsubject={},pdfkeywords={}}

\newcommand{\Kmax}{K_{\max}}
\newcommand{\abstain}{\bot}
\newtheorem{proposition}{Proposition}

\begin{document}
\maketitle

\begin{abstract}
Selecting less computation in a neural operator need not reduce inference
time. We study this gap using conformal calibration to choose Fourier
modes or channel width per input subject to a fixed error budget. A
triangle-inequality envelope bounds all candidates simultaneously; a
direct-score rule handles non-monotone candidate errors and, with an
input-independent score, calibrates a static baseline under the same
marginal guarantee. Under frozen-model and exchangeability assumptions,
both rules bound the probability of issuing a prediction outside the
budget and abstain when calibration is infeasible. On Burgers and Darcy
flow across three training seeds, mode selection reduces active spectral
support without a measurable latency reduction, because full-grid
transforms and channel mixing dominate the cost. Narrowing channels does
reduce Darcy forward cost, but reference-based selection must first
compute the dense prediction. An input-only binary width selector avoids
that pass and runs at $0.52$--$0.95\times$ dense latency; yet static
widths calibrated under the same rule are often as fast or faster, and
on a separately generated 1,000-example test set they generally achieve higher
coverage, while the selector falls below its 90\% target in more than
half of the settings. Adaptive computation in neural operators should therefore be judged by
achieved coverage and complete runtime against a matched static
comparator.
\end{abstract}

\section{Introduction}
\label{sec:intro}

A Fourier neural operator (FNO) uses the same spectral bandwidth and
channel width for every input \citep{li2021fno}, yet PDE instances can
differ substantially in complexity. This raises two questions: how much
of the trained operator does a particular input need, and can selecting
that amount reduce inference time while controlling error?

We study these questions with families obtained by truncating Fourier
modes in a trained FNO or by training shared weights at several channel
widths \citep{yu2019slimmable}. A split-conformal bound on the full
model's error combines with candidate disagreement to give an envelope
that holds for every candidate simultaneously. A direct-score variant
calibrates candidates' errors without assuming they decrease with cost.
Neither construction requires the error to be monotone in support or
width. The direct-score rule also selects one static width using an
input-independent score, giving a comparator with the same marginal
guarantee.

Our central question is whether a reduction in selected computation
survives complete runtime accounting. On Burgers, mode truncation leaves
full-grid transforms and pointwise channel mixing unchanged and yields
no measurable latency reduction. A Darcy sweep at batches 1 and 20 also
shows little change. Narrowing channels lowers batched Darcy forward
cost, but reference-based selection first computes the dense prediction.
We therefore study input-only routing as a concrete way to avoid that
cost. It beats dense inference in the measured configurations, yet
matched static models are faster at the tighter budgets. The comparison
identifies when the cost of adapting per input is repaid.

\paragraph{Contributions.}
We apply conformal calibration to spectral support and channel width,
with explicit calibration-level infeasibility and a static specialization
of the same direct-score rule. Our main empirical contribution is to
connect these guarantees to measured execution: we account for reference
passes, scoring, dispatch, and candidate inference. Across three training
seeds, we compare the resulting policies within each model family and
report their coverage on a separately generated Darcy test set. The
input-only routing case study exposes the tradeoff between static and
adaptive computation rather than establishing a general routing advantage.
Table~\ref{tab:practical-guidance} summarizes the constructions.

\begin{table}[H]
\centering
\caption{Roadmap: the four approaches studied in this paper. Runtime
findings apply only to the tested FNO and hardware.}
\label{tab:practical-guidance}
\small
\begin{tabular}{>{\raggedright\arraybackslash}p{0.22\linewidth}>{\raggedright\arraybackslash}p{0.32\linewidth}>{\raggedright\arraybackslash}p{0.36\linewidth}}
\toprule
Approach & Statistical role & Execution finding\\
\midrule
Reference-based spectral & Simultaneous error envelope for mode selection
& Little or no latency change in the tested Burgers and Darcy sweeps (Part I)\\[0.4em]
Reference-based width & Direct-score selection over nested widths
& Requires the dense pass, so no latency saving (Part II)\\[0.4em]
Binary TV selector & Input-only binary gate: widths $\{8,32\}$, scored by $\mathrm{TV}(x)$
& Full pipeline: $0.52$--$0.95\times$ dense latency on batched Darcy (Part II)\\[0.4em]
Static widths & Calibrated within each model family over $\{8,16,24,32\}$
& Runtime advantage depends on model family and tolerance (Part II)\\
\bottomrule
\end{tabular}
\end{table}

The formal assumptions, calibration-level abstention convention, and
data-split status are given in Sections~\ref{sec:method} and
\ref{sec:protocol}. The width routing case study uses a binary
$\{8,32\}$ gate, while its static comparator can choose any of the four
trained widths; this action-space difference is part of the scope of the
runtime comparison.

Section~\ref{sec:method} sets out the selection rules. Part~I
(Section~\ref{sec:results}) covers mode selection and its latency, and
Part~II (Section~\ref{sec:width}) covers channel width, from slimmable
training to input-only routing and the comparison with static models.

\section{Related work}
\label{sec:related}

\paragraph{Neural operators and adaptive computation.}
FNOs represent each layer using a truncated set of
Fourier modes, a choice that also underlies their approximation theory
\citep{li2021fno,kovachki2023neuraloperator,kovachki2021fnoapprox}. Several lines of work
modify the spectral computation or its cost, but do not select a mode
or channel budget for each input according to a prescribed error
tolerance. Adaptive FNOs use soft-thresholding to sparsify frequency
modes \citep{guibas2022afno}; incremental FNOs increase the number of
modes during training \citep{george2211incremental}; and factorized and
tensorized FNOs reduce the parameter cost of spectral weights
\citep{tran2023ffno,kossaifi2024mgtfno}. Outside operator learning,
elastic spectral state-space models can be truncated at inference time
to meet different computational budgets \citep{song2601elastic}.
Learned skip-block routing provides another way to adapt computation,
in transformer-based PDE models \citep{liu2025skipblock}. Meanwhile,
conformal methods for neural operators have focused on output intervals
and prediction regions in function space
\citep{chin2606conformalno,millard2509splitconformal,harris2025localoperator}.
We instead use conformal calibration to choose, for each input, how
many Fourier modes or channels to allocate while controlling the risk
of exceeding an error tolerance.

\paragraph{Conformal calibration and routing.}
\label{sec:related-routing}
\leavevmode
Our guarantees build on split-conformal prediction, risk-controlling
prediction sets, conformal risk control, and the Learn then Test (LTT)
framework for selecting predictive procedures under risk constraints
\citep{vovk2005conformal,lei2018distributionfree,bates2021rcps,
angelopoulos2022conformalrisk,angelopoulos2025learntest}. With an
indicator loss, expected risk is the marginal probability of a
violation. For an ordered width ladder, LTT fixed-sequence testing can
test each width's violation risk in a pre-specified sequence without
requiring wider widths to be accurate whenever a narrower width is
selected, but it guarantees, with probability at least $1-\alpha$ over
the calibration sample, that the selected width has risk at most
$\delta$, which requires an additional confidence level $\alpha$.
Split-conformal guarantees are instead marginal over the calibration draw.
For continuous scores, the coverage conditional on a calibration set
varies across calibration draws according to a Beta distribution.
Adjusting the quantile rank can control this conditional coverage with
high probability
\citep{vovk2013conditional,bian2023trainingconditional,angelopoulos2023gentle}.
Section~\ref{sec:width-selector} uses this calibration-conditional view
to interpret the observed coverage. Our matched static baseline instead
uses the direct-score rule, sharing the selector's marginal guarantee;
Section~\ref{sec:width-ceiling-bonferroni} explains why this matched
comparison is conservative.

Related work has also applied conformal calibration to early exits
against consistency with a full model, most closely in CALM
\citep{schuster2022calm}; this resembles our reference-disagreement
score, though CALM targets model consistency while we control error
against the target solution. Other work calibrates early exits
\citep{jazbec2405fastyetsafe}, reasoning budgets
\citep{wang2602conformalthinking}, and routing between cheap and
expensive models
\citep{uddin2603proactiverouting,uddin2026conformalrouting}. Without
conformal calibration, early exiting has also been cast as
classification with a reject option, calibrated to meet a fixed
computation budget \citep{valade2402eero}, and exit rules have
aggregated the confidence of several exit classifiers
\citep{bajpai2502beem}; these methods control cost, whereas our rules
control error against the target solution. Reachability analysis also
provides deterministic guarantees for cheaper models by bounding the
output error introduced by quantization \citep{cooke2304quantization}.

Like standard split-conformal guarantees, ours rely on exchangeability
between calibration and deployment data. Weighted conformal prediction
restores validity under covariate shift when the likelihood ratio
between test and calibration inputs is known or can be estimated
\citep{tibshirani2019covariateshift}; weighted quantiles bound the
coverage loss under more general departures \citep{barber2023beyond};
and adaptive conformal inference maintains long-run coverage online from
labelled feedback \citep{gibbs2021aci}. Group-wise (Mondrian)
calibration \citep{vovk2003mondrian} targets coverage within predefined
subpopulations. Our construction controls a marginal error event for
operator predictions, rather than risk conditional on the routed set.

\paragraph{Relationship to our compression study.}
Our earlier compression study \citep{ibrahim2026selection} evaluates
support selection in Fourier--wavelet dictionaries, resolution coupling,
residual correction, and transformed-channel compression, and already
emphasizes the difference between nominal sparsity and measured speed.
We share its canonical benchmark data, dense FNO training protocol, and
dense baseline checkpoints and errors. The present study adds marginally
calibrated per-input mode and width selection, its static
specialization, runtime accounting for reference passes and routing,
nested-width training, and the separately generated Darcy coverage
evaluation. A note in the code repository identifies the shared material;
the general observation that fewer operations need not imply lower latency
is background, not a new claim of this paper. 

\paragraph{Slimmable networks.}
\label{sec:related-slimmable}
Slimmable networks train a single checkpoint that runs at several widths
\citep{yu2019slimmable,yu2019universally}; we use the sandwich rule from
the latter. Nested representations also appear in Matryoshka learning
\citep{kusupati2022matryoshka}, and dynamic slimmable networks and
mixtures of slimmable experts learn gates that select the width
\citep{li2021dsnet,tastan2026mose}. Instead of learning a gate, we
choose the width with a calibrated rule, and we compare the nested
slices with independently trained models to measure what weight sharing
costs in accuracy.

\section{Benchmarks and experimental protocol}
\label{sec:benchmarks}

We use the Burgers and Darcy datasets publicly released by the authors
of the original FNO paper \citep{li2021fno}, split into 1000 training,
200 validation, and 200 test examples following the protocol of the
compression study \citep{ibrahim2026selection}.

\paragraph{Burgers.}
The model learns the one-shot map $u_0\mapsto u(\cdot,1)$ for
$u_t+uu_x=\nu u_{xx}$, with periodic boundaries, $\nu=0.01$, and
256 grid points. This is the released $\nu=0.01$ set; the Burgers
experiments in \citet{li2021fno} use the released $\nu=0.1$ set, which we
use only as a shifted test set (Appendix~\ref{app:shift}). The network
has width 56 and 16 trained modes.

\paragraph{Darcy.}
The model maps a piecewise-constant permeability field $a$ to the
solution of $-\nabla\!\cdot(a\nabla u)=f$ on a $141\times141$ grid.
It has width 32 and 12 modes per spectral axis. Its heterogeneous
coefficients contrast with the smoother Burgers fields.

Both networks append the grid coordinates to the input, apply a linear
lift, four spectral-plus-pointwise layers with GELU activations between
them, and a two-layer pointwise projection. Dense training uses 100
epochs of AdamW with learning rate $10^{-3}$, weight decay $10^{-4}$,
gradient clipping at 1.0, and StepLR with a factor of 0.5 every 25
epochs. The candidate mode counts are $K\in\{2,4,6,8,10,12,14,16\}$ for
Burgers and $K\in\{2,4,6,8,10,12\}$ for Darcy.

\subsection{Calibration, evaluation, and data reuse}
\label{sec:protocol}
Checkpoints are selected on the 200-example validation split. Table~\ref{tab:data-splits}
summarizes the calibration and evaluation sources; no calibration or
evaluation set in the main results was used for checkpoint selection.
Spectral Burgers results split the canonical test split 100/100, and
spectral Darcy results use the same calibration pool and new test set as
the width policies. The main width policies use a held-out
calibration pool of 624 examples from the second canonical piecewise-
constant Darcy data file and a separately generated 1,000-example
evaluation set. The score and budgets were chosen during earlier
development using the original test split; the new set evaluates those
locked choices on new examples, but does not remove that development reuse.

The new Darcy set uses the original FNO Gaussian-field generator with
$\alpha=2$, $\tau=3$, permeability levels $\{3,12\}$, and the finite-
difference solver at $421\times421$, followed by stride-3 subsampling to
$141\times141$. Runtime measurements use original canonical batches,
not the new coverage set. The generation history, including the discarded
pilot and overlap with an earlier 200-example run, is recorded in
Appendix~\ref{app:width-selector-robustness}.

\begin{table}[H]
\centering
\caption{Data sources and status of the reported evaluations.}
\label{tab:data-splits}
\small
\begin{tabular}{>{\raggedright\arraybackslash}p{0.18\linewidth}>{\raggedright\arraybackslash}p{0.19\linewidth}>{\raggedright\arraybackslash}p{0.17\linewidth}>{\raggedright\arraybackslash}p{0.16\linewidth}>{\raggedright\arraybackslash}p{0.12\linewidth}}
\toprule
Track & Model selection / development & Calibration & Evaluation & Status\\
\midrule
Spectral, Burgers & Validation ($n=200$) & Test half ($n=100$) & Other test half ($n=100$) & Held out from model selection\\
Spectral, Darcy & Validation ($n=200$) & Fresh canonical pool ($n=624$) & New Darcy set ($n=1000$) & Held out from model selection\\
Early width analyses & Validation and original test & Validation or fresh pool ($n=624$) & Original test ($n=200$) & Development\\
Main width coverage & Policies locked before generation & Fresh canonical pool ($n=624$) & New Darcy set ($n=1000$) & New examples; prior development reuse\\
Runtime & Locked width policies & Same fitted thresholds & Original canonical batches & Hardware benchmark\\
\bottomrule
\end{tabular}
\end{table}

\subsection{Error and cost conventions}
Coverage is the fraction of evaluation examples whose absolute error is
at most $\tau$, measured as the Euclidean norm of the unnormalized
output array. This absolute scale lets deployment disagreement and
reference error enter the same triangle bound; a relative error would
require the unknown target norm at deployment (Section~\ref{sec:certificate}).
It is computed on each benchmark's fixed grid, so it is
not a statement about coverage in function space across resolutions,
and the absolute budgets on the two benchmarks do not correspond to the
same relative accuracy. For an $m$-point output grid, the per-point RMS
budget is $\tau/\sqrt{m}$. Spectral selection uses
$\tau\in\{0.02,0.03,0.05,0.10\}$ on Burgers and
$\tau\in\{0.05,0.08,0.12,0.18\}$ on Darcy; width selection uses
$\tau\in\{0.020,0.024,0.028\}$ on Darcy. We use different Darcy budget ranges for spectral and width experiments
because the former uses a reference-error envelope, while the latter
calibrates candidate errors directly; the ranges yield informative,
feasible operating points for those distinct rules. For scale, the
spectral reference-error quantiles $\hat q_{0.10}$ are $0.0059$--$0.0071$ on
Burgers and $0.0176$--$0.0184$ on Darcy, and at the loosest width budget
the selected widths have 1.5--1.8 times the mean dense error
(Appendix~\ref{app:width-reference}).
Model-accuracy tables report relative $L^2$ error, as does the preliminary oracle diagnostic in
Appendix~\ref{sec:oracle}, which does not set the calibration budgets.

Timing starts with raw input on the CPU and normalized batches on the
GPU; loading and normalization are excluded. Complete routing includes
CPU scoring, threshold transfer, dispatch, and recombination. Nested
static timings include in-call slicing; independent static timings do
not. The support fraction is the mean selected $K$ divided by $\Kmax$. For
width, we report three cost measures: a proxy that weights isolated-width
latencies by routing fractions, dispatch-only timings with scores
precomputed, and the complete pipeline including scoring. The complete
pipeline is the headline runtime. All timings use an RTX A6000 with 20
warm-up runs and 100 synchronized repeats. When calibration is
infeasible, we report it as such and assign no coverage to the
abstaining policy.

\section{Selection rules}
\label{sec:method}

Both rules use a fixed absolute error budget $\tau$ and a risk level
$\delta$. We state them with the spectral support $K$ as the action; for
width, replace $K$ with $w$. Table~\ref{tab:notation} lists the
notation.

\begin{table}[H]
\centering
\caption{Notation for calibrated selection. All absolute errors use
the benchmark's fixed output grid.}
\label{tab:notation}
\small
\begin{tabular}{>{\raggedright\arraybackslash}p{0.25\linewidth}>{\raggedright\arraybackslash}p{0.67\linewidth}}
\toprule
Symbol & Meaning\\
\midrule
$x,\ u^\star(x)$ & Input field and target solution\\
$K,\ \Kmax;\ w,\ w_{\max}$ & Candidate and full spectral support; candidate and full width\\
$u_K(x),\ u_w(x)$ & Predictions at the selected support or width\\
$e_K,\ e_w,\ e_{\max}$ & Absolute prediction errors; $e_{\max}$ is the full-reference error\\
$\eta(x,K),\ \eta(x,w)$ & Absolute disagreement with the corresponding full prediction\\
$\tau,\ \delta$ & Fixed absolute error budget and allowed marginal error-event probability\\
$n,\ \hat q_\delta$ & Calibration sample size and $(1-\delta)$ upper conformal quantile of reference error\\
$h(x,K),\ s(x)$ & Deployment score and labelled direct calibration score\\
$\hat\lambda$ & Calibrated lower threshold for direct-score selection\\
$\hat K(x),\ \hat w(x),\ \abstain$ & Selected action; $\bot$ means calibration-level infeasibility, so no deployment prediction is issued for that fitted rule\\
$\mathrm{TV}(x)$ & Mean absolute input finite difference across spatial axes\\
\bottomrule
\end{tabular}
\end{table}

\subsection{A nested family from one trained operator}
To lower the spectral support, we keep only a prefix of each layer's
stored Fourier multipliers. At full support this is the ordinary forward
pass, so one trained model gives a nested family
$u_{K_1},\ldots,u_{\Kmax}$ without retraining. The weights are shared,
but each candidate is a full nonlinear forward pass rather than a
projection of the dense output. Our tests confirm exact reproduction at
full support, reversible slicing, and correct output shapes.

\Needspace{12\baselineskip}
\subsection{The certificate}
\label{sec:certificate}
For an input $x$, the candidate and reference errors are
\[
e_K(x)=\lVert u_K(x)-u^\star(x)\rVert_2,\qquad
e_{\max}(x)=\lVert u_{\Kmax}(x)-u^\star(x)\rVert_2,
\]
and the disagreement, which can be computed at deployment, is
\[
\eta(x,K)=\lVert u_K(x)-u_{\Kmax}(x)\rVert_2.
\]
Here $\lVert\cdot\rVert_2$ is the Euclidean norm of the flattened,
unnormalized output array on the benchmark's fixed grid, so errors and
disagreements have the same units. The tolerance is absolute because a
relative one would need the norm of the unknown target. The reference
prediction is computed once per input and reused for every candidate.

\paragraph{Step 1: a deterministic bound.}
For any weights and every candidate,
\begin{equation}
e_K(x)\le\eta(x,K)+e_{\max}(x),
\label{eq:triangle}
\end{equation}
by the triangle inequality applied to
$u_K-u^\star=(u_K-u_{\Kmax})+(u_{\Kmax}-u^\star)$.
The collected error grids show no violations beyond numerical
tolerance.

\paragraph{Step 2: a simultaneous marginal envelope.}
Suppose the model family and score design are fixed before calibration,
using data independent of the calibration set, and that the $n$
labelled calibration examples and a new example are exchangeable. Let $r=\lceil(1-\delta)(n+1)\rceil$. For calibration values
$z_1,\ldots,z_n$, define the upper conformal quantile
$Q_{1-\delta}(z_1,\ldots,z_n)=z_{(r)}$ when $r\le n$, and
$+\infty$ otherwise, where $z_{(r)}$ is the $r$-th order statistic.
Set $\hat q_\delta=Q_{1-\delta}(e_{\max}(x_1),\ldots,e_{\max}(x_n))$. The standard quantile lemma gives
$\Pr(e_{\max}(x_{\mathrm{new}})\le\hat q_\delta)\ge1-\delta$
\citep{vovk2005conformal,lei2018distributionfree}. Since
Eq.~\eqref{eq:triangle} holds for every $K$,
\begin{proposition}[Simultaneous reference envelope]
Under the frozen-model and exchangeability assumptions above,
\begin{equation}
\Pr\!\left(\forall K:\ e_K(x_{\mathrm{new}})
 \le\eta(x_{\mathrm{new}},K)+\hat q_\delta\right)\ge1-\delta.
\label{eq:envelope}
\end{equation}
A single event on the reference error covers every candidate at once,
so choosing among the envelopes needs no union bound over $K$.
\end{proposition}
\begin{proof}
The conformal quantile lemma gives the event
$e_{\max}(x_{\mathrm{new}})\le\hat q_\delta$ with probability at
least $1-\delta$. On this single event, the pointwise triangle
inequality holds for every candidate $K$ simultaneously.
\end{proof}

\paragraph{Step 3: selection with explicit infeasibility.}
For a fixed finite absolute tolerance $\tau$, define
\begin{equation}
\hat K(x)=
\begin{cases}
\min\{K:\eta(x,K)\le\tau-\hat q_\delta\}, & \hat q_\delta\le\tau,\\
\abstain, & \hat q_\delta>\tau .
\end{cases}
\label{eq:stopping}
\end{equation}
The candidate set includes $\Kmax$, whose disagreement is zero, so the
minimum exists whenever $\hat q_\delta\le\tau$. Otherwise the rule
returns $\abstain$ and issues no prediction; this means only that this
construction found no eligible action. Equation~\eqref{eq:envelope}
implies
\begin{equation}
\Pr\big(\hat K(x_{\mathrm{new}})\ne\abstain,\
 e_{\hat K(x_{\mathrm{new}})}(x_{\mathrm{new}})>\tau\big)\le\delta.
\label{eq:guarantee}
\end{equation}
The probability is over the calibration set and the new example, with
the model family held fixed. If the rule always issues a prediction,
Eq.~\eqref{eq:guarantee} reduces to marginal coverage at level
$1-\delta$. Equivalently, with probability at least $1-\delta$, the fitted rule
either abstains or issues a prediction with error at most $\tau$. This
is an unconditional joint statement over calibration and deployment
data, not a bound conditional on calibration being feasible; conditional
miscoverage can be as large as $\delta/\Pr(\text{feasible})$. The
result does not bound how often calibration is feasible.
Appendix~\ref{app:safeguards} gives numerical conventions.

\subsection{A direct-score rule without monotone action errors}
\label{sec:robust-score}
The triangle envelope can be conservative. The direct-score rule instead
calibrates a threshold on the candidates' own errors, under the same
assumptions and with the same guarantee. The spectral results in Section~\ref{sec:results}
report both rules; the width results use this direct-score rule.

Because FNO layers are nonlinear, neither the errors nor the
disagreements need to decrease monotonically as support grows. The score
therefore takes every inaccurate candidate into account, rather than
only the score at the smallest accurate support;
Appendix~\ref{app:robust-score-unsound} shows how that simpler rule
fails.

\paragraph{A conservative score.}
For any frozen, finite-valued deployment score $h(x,K)$, define
\begin{equation}
s(x)=
\begin{cases}
-\infty, & e_{\max}(x)>\tau,\\
\min\{h(x,K):K<\Kmax,\ e_K(x)>\tau\}, & e_{\max}(x)\le\tau ,
\end{cases}
\label{eq:robust-score}
\end{equation}
where $K$ ranges over the candidate set and the minimum of an empty set
is $+\infty$. Setting $s(x)=-\infty$ when the full model itself misses
the budget is conservative. Such calibration examples remain in the
quantile calculation and consume the allowed $\delta$ failure budget;
if there are too many, calibration becomes infeasible. They are outside
the conformal proof event even if a smaller candidate happens to be
accurate, so the rule does not use the convention to discount hard cases.

Let $\hat\lambda=-Q_{1-\delta}(-s_1,\ldots,-s_n)$, using the
upper conformal quantile defined in Section~\ref{sec:certificate}. The
score quantile gives $\Pr(s(x_{\mathrm{new}})\ge\hat\lambda)\ge1-\delta$.
The deployed action is
\begin{equation}
\hat K(x)=\begin{cases}
\abstain, & \hat\lambda=-\infty,\\
\min\bigl(\{K<\Kmax:h(x,K)<\hat\lambda\}\cup\{\Kmax\}\bigr), & \hat\lambda>-\infty.
\end{cases}
\label{eq:direct-selector}
\end{equation}
\begin{proposition}[Direct-score selection]
Under the same frozen-model and exchangeability assumptions, the policy
in Eq.~\eqref{eq:direct-selector} satisfies
\[
\Pr\bigl(\hat K(x_{\mathrm{new}})\ne\abstain,\
 e_{\hat K(x_{\mathrm{new}})}(x_{\mathrm{new}})>\tau\bigr)\le\delta.
\]
\end{proposition}
\begin{proof}
On the event $s(x_{\mathrm{new}})\ge\hat\lambda>-\infty$, the fallback
is accurate and every inaccurate candidate has
$h(x_{\mathrm{new}},K)\ge s(x_{\mathrm{new}})\ge\hat\lambda$, so it
cannot be selected. The conformal quantile event holds with probability
at least $1-\delta$; strict eligibility excludes ties.
\end{proof}

Taking $h=\eta$ gives the reference-based variant. Giving every
non-maximal width the same input statistic as its score gives the binary
rule of Section~\ref{sec:width-selector}. That rule can be more
conservative than calibrating only the two widths it actually uses,
since errors at the unused intermediate widths still enter the score; we
keep it because it matches the recorded experiments.

For an example on which the full width and all non-maximal widths
are accurate, $s(x)=+\infty$. This is an individual calibration score;
the deployed policy depends on the calibrated quantile across all
examples. If that threshold is $+\infty$, every finite deployment score
is eligible and the rule always chooses the narrowest candidate.

Section~\ref{sec:results} compares both rules on spectral support.

\section{Part I: adaptive spectral support}
\label{sec:results}

This part applies both rules to the number of Fourier modes in
the original dense models. A preliminary diagnostic that uses the labels
shows that the support needed varies across Burgers inputs
(Appendix~\ref{sec:oracle}). Here we evaluate the rule that can be run
at deployment: its latency, the support it selects, and how that support
relates to input complexity.

\subsection{Mode-only latency}
\label{sec:g4}
\begin{figure}[!htbp]
\centering
\includegraphics[width=0.62\linewidth]{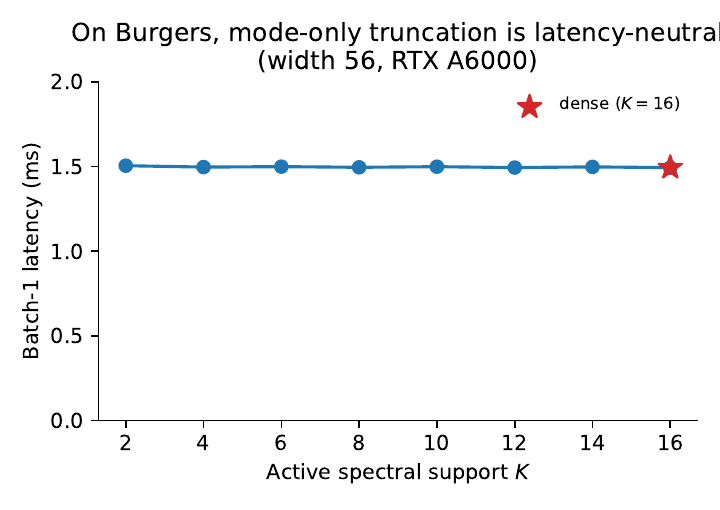}
\caption{Batch-1 latency across spectral support, Burgers, width 56,
RTX A6000. The axis starts at zero; changing $K$ leaves elapsed time
effectively unchanged, with recorded means of 1.4934--1.5047\,ms.}
\label{fig:latency}
\end{figure}

Changing $K$ gives no measurable latency reduction on Burgers across the
tested mode counts. Figure~\ref{fig:latency} shows the batch-1
measurements; the sweep also covered batch 20.

We also timed a trained Darcy FNO at batch sizes 1 and 20
(Table~\ref{tab:darcy-mode-latency}); these were the only Darcy batch
sizes measured for mode truncation. At batch 20, reducing $K$ from
12 to 2 lowered measured latency by 0.7\%; at batch 1 the difference
was 0.3\%. These reductions are small on this checkpoint and hardware.

\begin{table}[H]
\centering
\caption{Mode-truncation latency on Darcy, standard width-32 FNO, seed
42, 141$\times$141 grid, RTX A6000. Means and standard deviations over
100 synchronized repeats after 20 warm-ups, in milliseconds.}
\label{tab:darcy-mode-latency}
\small
\begin{tabular}{rrr}
\toprule
$K$ & Batch 1 & Batch 20\\
\midrule
2  & 1.816 $\pm$ 0.022 & 11.032 $\pm$ 0.029\\
4  & 1.808 $\pm$ 0.022 & 11.045 $\pm$ 0.047\\
6  & 1.838 $\pm$ 0.262 & 11.054 $\pm$ 0.042\\
8  & 1.812 $\pm$ 0.021 & 11.066 $\pm$ 0.045\\
10 & 1.808 $\pm$ 0.027 & 11.097 $\pm$ 0.045\\
12 & 1.821 $\pm$ 0.023 & 11.114 $\pm$ 0.066\\
\bottomrule
\end{tabular}
\end{table}

Most of the work does not depend on $K$. Pointwise channel mixing costs
$O(w^2N)$ and transforms still run on the full grid. The spectral
contraction scales as $O(K^d w^2)$ for $d$ spatial dimensions: linearly
in $K$ for Burgers and quadratically for the two-dimensional Darcy
implementation. Fixed transform plans, launch overhead, and full-grid
spectral tensor handling can dominate these contractions at the tested
resolutions. We did not measure Darcy mode truncation at batch 64 or 128,
so its latency result should not be extrapolated to the headline routing
batches.
Reference-based selection also has to compute the dense prediction and
any candidates it probes. For the spectral axis we therefore report
selected support as well as these small timing changes; the larger
runtime gains in Part~II come from reducing channel width.
\FloatBarrier

\subsection{Support selection on Burgers and Darcy}
\label{sec:seeds}
Tables~\ref{tab:certified-burgers} and~\ref{tab:certified-darcy} report
both rules for all three training seeds on data not used for checkpoint
selection (Section~\ref{sec:protocol}). The same code runs on both
benchmarks without any PDE-specific residual, and the triangle bound
(Eq.~\eqref{eq:triangle}) holds on every collected instance--support
pair.

\begin{table}[H]
\centering
\caption{Spectral selection on Burgers, calibrated and evaluated on
disjoint 100-example halves of the canonical test split. Per-seed values
are listed as seeds 42 / 43 / 44; $\hat q_{0.10}=0.0059$--$0.0071$.}
\label{tab:certified-burgers}
\small
\resizebox{\linewidth}{!}{%
\begin{tabular}{rrcccc}
\toprule
& & \multicolumn{2}{c}{Triangle rule} & \multicolumn{2}{c}{Direct-score rule}\\
\cmidrule(lr){3-4}\cmidrule(lr){5-6}
$\tau$ & $\delta$ & Coverage & Support fraction (\%) & Coverage (\%) & Support fraction (\%)\\
\midrule
0.02 & 0.10 & 99.0--100.0\% & 52.2 / 54.2 / 55.0 & 92.0 / 88.0 / 94.0 & 46.0 / 45.1 / 49.8\\
0.03 & 0.10 & 100.0\% & 44.2 / 44.9 / 47.2 & 95.0 / 94.0 / 91.0 & 41.2 / 41.5 / 43.2\\
0.05 & 0.10 & 100.0\% & 39.2 / 39.4 / 40.8 & 83.0 / 85.0 / 83.0 & 35.6 / 36.8 / 38.1\\
0.10 & 0.10 & 100.0\% & 30.6 / 31.4 / 32.0 & 89.0 / 89.0 / 88.0 & 28.9 / 29.1 / 29.8\\
0.03 & 0.05 & 100.0\% & 45.2 / 45.5 / 48.0 & 97.0 / 99.0 / 95.0 & 41.5 / 42.1 / 43.8\\
\bottomrule
\end{tabular}}
\end{table}

\begin{table}[H]
\centering
\caption{Spectral selection on Darcy, calibrated on the fresh canonical
pool ($n=624$) and evaluated on the new Darcy test set ($n=1000$),
$\delta=0.10$. Per-seed values are listed as seeds 42 / 43 / 44;
$\hat q_{0.10}=0.0176$--$0.0184$.}
\label{tab:certified-darcy}
\small
\resizebox{\linewidth}{!}{%
\begin{tabular}{rcccc}
\toprule
& \multicolumn{2}{c}{Triangle rule} & \multicolumn{2}{c}{Direct-score rule}\\
\cmidrule(lr){2-3}\cmidrule(lr){4-5}
$\tau$ & Coverage & Support fraction (\%) & Coverage (\%) & Support fraction (\%)\\
\midrule
0.05 & 99.6\% & 99.7 / 99.9 / 99.9 & 85.4 / 88.0 / 86.7 & 94.9 / 97.0 / 96.2\\
0.08 & 99.7--100.0\% & 93.0 / 97.4 / 95.4 & 89.1 / 90.1 / 87.7 & 84.5 / 90.5 / 86.9\\
0.12 & 99.6--99.7\% & 71.1 / 81.1 / 76.0 & 90.8 / 93.1 / 90.2 & 59.8 / 69.7 / 64.0\\
0.18 & 99.5--99.8\% & 37.6 / 45.0 / 40.1 & 92.0 / 91.9 / 88.4 & 30.2 / 35.8 / 31.3\\
\bottomrule
\end{tabular}}
\end{table}

The triangle rule covers 99--100\% of evaluation examples throughout.
On Burgers it selects 31--55\% of the trained modes; on Darcy it keeps
almost all modes at $\tau=0.05$--0.08 and reduces support only at the
looser budgets. The contrast between the benchmarks holds across seeds.
The direct-score rule trades the triangle rule's slack for smaller
supports, by 1.8--12.0 percentage points (mean 5.7). Its coverage is
83--99\% on Burgers, where $n=100$ makes each value imprecise, and
85.4--93.1\% on Darcy, below 90\% in 6 of 12 cells. At $\tau=0.05$, the
Darcy shortfalls for seeds 42 and 44 are larger than random re-splits of
the pooled data usually produce (Appendix~\ref{app:spectral-resplit}). The dense models' relative
$L^2$ test errors for seeds 42, 43, and 44 are 0.002458, 0.002586, and
0.002522 on Burgers and 0.012107, 0.012116, and 0.012040 on Darcy.

\subsection{Input complexity and selected support}
\label{sec:complexity}
\begin{figure}[!htbp]
\centering
\includegraphics[width=\linewidth]{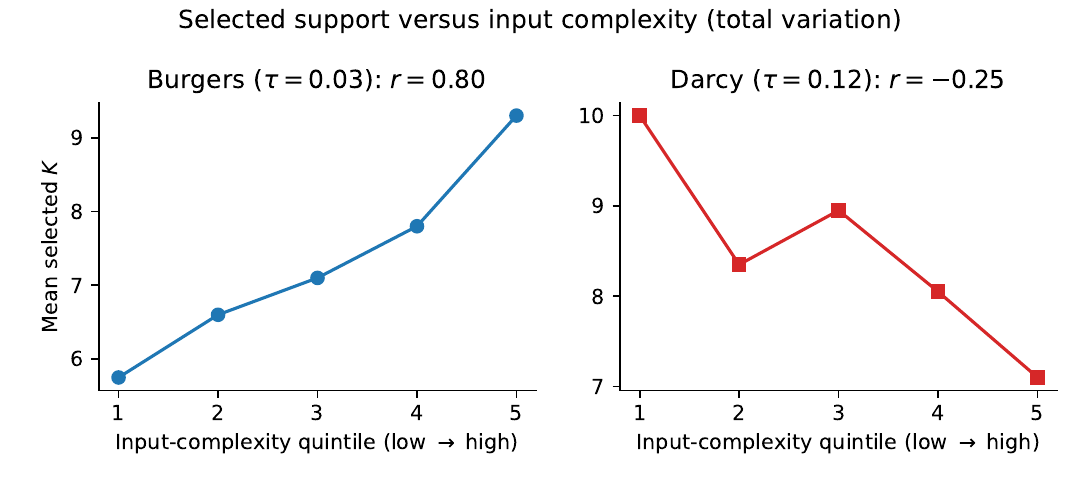}
\caption{Mean selected spectral support by input total-variation
quintile. Burgers shows a monotonic increase; Darcy does not.}
\label{fig:complexity}
\end{figure}

As a simple measure of input complexity we use total variation, the
mean absolute finite difference of the input across spatial axes.These
descriptive correlations apply the triangle rule, with $\hat q$ from one
half of the 200-example validation split, to all 200 validation examples. On
Burgers at $\tau=0.03$, it correlates with selected support at
$r=0.80$, and mean $\hat K$ rises from 5.75 in the smoothest quintile
to 9.30 in the roughest (Figure~\ref{fig:complexity}).

Darcy behaves differently: the correlations are $r=-0.08$, $-0.25$, and
$-0.37$ at $\tau=0.08$, 0.12, and 0.18. Using the total variation of the
target solution instead gives correlations of only 0.02--0.11. At tight
tolerances most selected supports sit at $\Kmax$;
Appendix~\ref{app:range-restriction} looks at the remaining inputs.
Input total variation therefore predicts spectral selection on Burgers
but does not give a common ordering of difficulty across the two PDEs.

\paragraph{Comparison with width.}
\label{sec:complexity-reconcile}
On Darcy, total variation is positively correlated with the minimum
accurate width ($r=+0.22$ to $+0.52$; Part~II), even though it is
negatively correlated with selected spectral support at loose
tolerances. Width and modes are different targets. Elliptic smoothing
and the capacity of channel mixing are plausible explanations, but our
experiments do not isolate a mechanism. A complexity signal has to be
checked against the specific choice it is meant to predict.
\FloatBarrier

\section{Part II: channel width and a routing case study}
\label{sec:width}

Narrowing channels reduces pointwise channel mixing and the number of
full-grid transforms. We first measure this effect and account for the
cost of reference-based selection. We then use input-only routing as a
case study of adaptive execution, comparing it with a static width
calibrated under the same marginal guarantee.

\subsection{A nested family with width-dependent runtime}
\label{sec:width-method}
\label{sec:width-latency}
We train one Darcy FNO at widths $\{8,16,24,32\}$ with the sandwich rule
\citep{yu2019universally}, summing the four losses before each backward
pass. On the original test split, its full-width error differs from that of independent
dense training by $-0.07\%$, $+12.85\%$, and $-5.07\%$ for the three
seeds. Weight sharing costs most at width 8, whose error is on average
19.3\% higher than that of independently trained width-8 models; at the
other widths the mean gap is 1.5--5.5\%. Appendix~\ref{app:width-training}
gives the full accuracy results.

\begin{figure}[!htbp]
\centering
\includegraphics[width=0.62\linewidth]{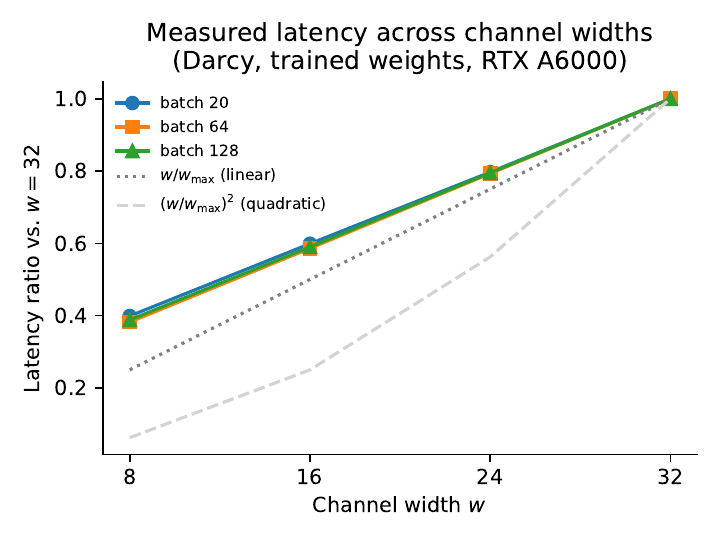}
\caption{Latency by channel width, trained Darcy checkpoint, RTX A6000.
Each ratio uses width 32 at the same batch size as its reference.
Linear and quadratic curves provide operation-scaling comparisons.}
\label{fig:width-latency}
\end{figure}

An initial sweep over widths $\{8,16,24,32,40,48,56\}$ and batch sizes
$\{1,20,64,128\}$ found useful latency reductions from narrowing on
Darcy at batch sizes of 20 or more, but not on Burgers or on Darcy at
batch 1. We then timed the trained nested checkpoint, with weight
slicing inside the timed call. On batched Darcy, width 8 takes
$0.38$--$0.40\times$ the width-32 latency
(Figure~\ref{fig:width-latency}).

The ratios scale closer to linearly than quadratically with width.
This gap between narrow and wide inference is what adaptive execution
can exploit.

\subsection{Reference-based selection pays for dense inference}
The disagreement carries over directly as
$\eta(x,w)=\lVert u_w(x)-u_{32}(x)\rVert_2$, and it tracks the error
closely in the recorded grids (Spearman correlation 0.74--0.91).
Disagreement is monotone in width, but the true errors are non-monotone
on 21--32\% of examples. The direct-score rule of
Section~\ref{sec:robust-score} does not rely on either ordering.

The problem is cost. Computing $\eta(x,w)$ needs $u_{32}(x)$ for every
input, so the dense pass is paid on every input, on top of the candidate
passes and the selection itself. Reference-based selection is still useful for checking narrower widths
against the full model, but a speedup needs a routing signal that is
available before the dense prediction. Appendix~\ref{app:width-reference}
reports the error of the selected widths relative to the dense model and
compares against a label-informed reference policy.

\subsection{Case study: input-only selection}
\label{sec:width-selector}
\label{sec:width-runtime}
The total variation of the input coefficient field is a moderate
predictor of the minimum accurate width, with $r=+0.22$ to $+0.52$
across the seed--budget combinations we examined. We also tried
phase-area fraction, connectivity/percolation, and spectral tail energy,
but none gave a comparably useful signal. The choice of score is one of
the development decisions described in Section~\ref{sec:protocol}.

We use total variation as the deployment score by setting
$h(x,w)=\mathrm{TV}(x)$ for every non-maximal width in
Eq.~\eqref{eq:robust-score}. Since all non-maximal widths then share one
score, the policy chooses only between widths 8 and 32, even though the
model family also contains widths 16 and 24. The score in the recorded
calibration rule still includes failures at those unused widths; this is
a conservative feature of the recorded binary experiment. For a binary
action set $\{8,32\}$ alone, the score simplifies to
$s_{\mathrm{bin}}(x)=-\infty$ if $e_{32}(x)>\tau$, $\mathrm{TV}(x)$
if $e_{32}(x)\le\tau<e_8(x)$, and $+\infty$ if both widths pass.
The deployed rule then chooses either width 8 or the full width:
\[
\hat w(x)=
\begin{cases}
\abstain, & \hat\lambda=-\infty,\\
8, & \mathrm{TV}(x)<\hat\lambda,\\
32, & \mathrm{TV}(x)\ge\hat\lambda .
\end{cases}
\]
Calibration uses the fresh pool of $n=624$ examples and the three
budgets from Section~\ref{sec:protocol}. At deployment, each input goes
to exactly one width, so the two forward passes run on disjoint
sub-batches.

\paragraph{Measured execution.}
The complete pipeline (CPU NumPy scoring, threshold transfer, GPU
sub-batch dispatch, and recombination) takes $0.520$--$0.948\times$ the
dense latency across all 24 feasible combinations of seed, budget, and
batch size. Dispatch alone takes $0.41$--$0.84\times$
dense, and CPU scoring about 10--11\% of a dense pass. The isolated GPU
scorer suggests a lower additive cost, but a combined GPU-scoring
pipeline was not measured. Batch-64 timings use original canonical
batches; at $\tau=0.024$, their local width-8 fractions are 7--10 points
below the fractions on the new coverage set. These timings therefore do
not reproduce the new test-set routing composition. Appendix~\ref{sec:tv-cost}
gives component timings and the per-batch breakdown.

\paragraph{Prediction coverage and routing.}
Table~\ref{tab:matched-static} reports coverage, routing fractions, and
full-pipeline timing beside the matched static policies. Across the
feasible nested rows, coverage is 87.3--90.5\% against a 90\% target,
and 33.7--96.3\% of inputs go to width 8; seed 43 at $\tau=0.020$ is
infeasible because calibration gives $\hat\lambda=-\infty$. Coverage
falls below target in six of eight feasible nested cells and four of
nine independent cells. At $n=1000$, the binomial standard error near
90\% is about 0.95 percentage points. The pointwise exact 95\% intervals
lie entirely below 90\% only for nested seed 42 at $\tau=0.024$
(85.1--89.3\%) and seed 43 at $\tau=0.028$ (85.4--89.6\%); the other six
nested intervals include 90\%, and the three seed-44 independent
intervals lie entirely above it. The cells share examples, so these are
not simultaneous intervals. The marginal theorem does not ensure 90\% coverage conditional
on a particular calibration set or in every finite test sample. These
results therefore leave uncertainty about the achieved risk of the fitted
policies; they do not establish that every operating point meets its target.

Calibration variability alone is substantial at this sample size. For a
continuous, tie-free score, the probability of the conformal event
conditional on the calibration set follows a
$\mathrm{Beta}(r,n+1-r)$ distribution, here $\mathrm{Beta}(563,62)$
for $n=624$ and $\delta=0.10$: mean 90.1\%, standard deviation 1.2
percentage points, and central 95\% range 87.6--92.3\%. Ties, the
infinite score values, and the fact that achieved coverage contains
the conformal event can only make conditional coverage stochastically
larger. Adding test-set sampling at $n=1000$ gives a combined standard
deviation of about 1.5 points; under this boundary case, observing at
most 873 or 876 covered examples has probability of about 4\% and 6\%.
The observed shortfalls are therefore of the size expected from a single
calibration draw, although this does not establish their cause, and the
cells share one calibration pool per seed, so it is not a multiplicity
correction.
Appendix~\ref{app:width-selector-robustness} distinguishes this new test
from the earlier resampling diagnostics.

\subsection{Comparison with static models}
\label{sec:width-ceiling}
A static narrow model is the relevant comparator for deciding whether
per-input adaptation is useful. We compare within each model family:
nested slices against the nested selector, and independently trained
widths against their corresponding selector.

\subsubsection{A static baseline with the same marginal guarantee}
\label{sec:width-ceiling-bonferroni}
The descriptive per-width quantile screen (Table~\ref{tab:width-ceiling},
Appendix~\ref{app:static-screen}) inspects four quantiles and reports
the cheapest width, so its per-width $90\%$ bounds do not give the
selected width a $90\%$ guarantee; Bonferroni at $\delta/4$ would be
valid but unnecessarily conservative. Instead, we calibrate a static action with the direct-score rule from
Eq.~\eqref{eq:robust-score}. Set $h(x,w)=g(w)$ for a fixed strictly
decreasing function, here $g(w)=-w$. Every input then receives the same
width. The score is $s(x)=g(w_{\mathrm{last}})$, where
$w_{\mathrm{last}}$ is the largest non-maximal width whose error exceeds
$\tau$, and is $-\infty$ when width 32 itself exceeds $\tau$, using
the same convention as the adaptive selector. If the full width and all
non-maximal widths are accurate, $s(x)=+\infty$. The same lower conformal
quantile therefore controls the same event,
$\Pr(\text{issue a prediction and its error exceeds }\tau)\le\delta$,
without a union bound over widths. We calibrate nested and independently
trained model families separately; no minimum is taken across families.

This cost-ordered static rule controls a stronger event than the marginal
accuracy of its one selected width: its score screens errors at wider
widths in the suffix of the ladder and abstains on the same calibration
examples for which the full width fails. That coupling makes it a matched
comparison under the same direct-score construction, but can be
conservative for independent models whose width errors are not nested.
In particular, for the selected static width $\hat w$, the conformal
event implies
$\Pr(\hat w\ne\abstain,\ \forall w'\ge\hat w:\ e_{w'}(x_{\mathrm{new}})\le\tau)\ge1-\delta$,
which is stronger than controlling only $e_{\hat w}(x_{\mathrm{new}})$.
Fixed-sequence LTT testing can test individual width risks in a
pre-specified order without requiring a suffix of widths to pass, but it
provides high-probability rather than marginal control
(Section~\ref{sec:related-routing}). We did not run that alternative on
this calibration pool, so the static results below are not claimed to be
the best possible static policy.

Table~\ref{tab:matched-static} reports each resulting fixed width and
empirical test coverage beside the corresponding selector. It compares
the selector's complete batch-64 runtime with the isolated forward-pass
time of its matched static model. Nested static timings include in-call
weight slicing. We did not time slicing separately, but the nested and
independent static ratios agree to three decimals at widths 16 and 24,
which suggests the slicing cost is negligible at batch 64. Feasibility agrees within each model
family because their scores assign $-\infty$ to exactly the same
calibration examples: those whose full-width error exceeds $\tau$.

\begin{table}[H]
\centering
\caption{Family-matched comparison at $\delta=0.10$ on Darcy. Each
static width is selected by the direct-score rule with the
input-independent score $h(x,w)=-w$. Coverage and routing fractions (w8)
are on the 1,000-example new test set; times are batch-64 latencies of
the same fitted policies on separate original canonical batches, on RTX
A6000 GPUs (the independent family on a single node), normalized by each
family's own width-32 time. Selector times include CPU scoring, dispatch, and
recombination; static times are single forward passes, with in-call
weight slicing for nested models only. Infeasible means the calibration
rule abstains. The two independent-family infinite-threshold rows are
constant width-8 policies whose generic pipeline overhead can be
bypassed.}
\label{tab:matched-static}
\scriptsize
\resizebox{\linewidth}{!}{%
\begin{tabular}{crcccc}
\toprule
Seed & $\tau$ & \shortstack{Nested routing\\coverage / w8 / time} & \shortstack{Nested static\\width/coverage/time} & \shortstack{Independent routing\\coverage / w8 / time} & \shortstack{Independent static\\width/coverage/time}\\
\midrule
42 & 0.020 & 89.6\% / 34.6\% / 0.884$\times$ & 24 / 92.4\% / 0.792$\times$ & 89.9\% / 43.6\% / 0.786$\times$ & 24 / 93.2\% / 0.792$\times$\\
42 & 0.024 & 87.3\% / 73.5\% / 0.708$\times$ & 16 / 95.6\% / 0.586$\times$ & 89.2\% / 83.9\% / 0.589$\times$ & 16 / 96.6\% / 0.586$\times$\\
42 & 0.028 & 90.5\% / 94.2\% / 0.536$\times$ & 16 / 97.3\% / 0.586$\times$ & 91.5\% / 100.0\% / 0.466$\times$ & 8 / 91.5\% / 0.384$\times$\\
43 & 0.020 & infeas. & infeas. & 89.7\% / 38.3\% / 0.834$\times$ & 24 / 95.2\% / 0.792$\times$\\
43 & 0.024 & 89.3\% / 61.8\% / 0.775$\times$ & 16 / 91.6\% / 0.586$\times$ & 89.5\% / 73.7\% / 0.686$\times$ & 16 / 93.9\% / 0.586$\times$\\
43 & 0.028 & 87.6\% / 96.3\% / 0.520$\times$ & 16 / 95.7\% / 0.586$\times$ & 90.7\% / 98.5\% / 0.479$\times$ & 16 / 96.4\% / 0.586$\times$\\
44 & 0.020 & 89.9\% / 33.7\% / 0.887$\times$ & 24 / 93.8\% / 0.792$\times$ & 92.8\% / 52.9\% / 0.774$\times$ & 32 / 94.0\% / 1.000$\times$\\
44 & 0.024 & 89.7\% / 66.2\% / 0.771$\times$ & 16 / 95.8\% / 0.586$\times$ & 93.8\% / 99.6\% / 0.467$\times$ & 16 / 92.1\% / 0.586$\times$\\
44 & 0.028 & 90.2\% / 86.9\% / 0.607$\times$ & 16 / 97.7\% / 0.586$\times$ & 96.6\% / 100.0\% / 0.467$\times$ & 8 / 96.6\% / 0.383$\times$\\
\bottomrule
\end{tabular}}
\end{table}

Matched static coverage exceeds nested-selector coverage in all eight
feasible nested cells. The independent static comparator has higher
coverage in six of nine cells and equal coverage in the two cells with an
infinite routing threshold, where both policies use width 8. At seed 44
and $\tau=0.024$, its coverage is 1.7 points lower than routing.
The higher static coverage is consistent with two structural features
rather than with better calibration. First, the static rule controls the
stronger suffix event described above. Second, the static action moves on
a coarse four-width ladder: when width 8 does not clear the threshold,
the rule must step to width 16 or 24, whose coverage at that budget can be
well above 90\%. The routing threshold is instead set on a continuous
score and can sit close to the calibration quantile.

\begin{figure}[!htbp]
\centering
\includegraphics[width=0.72\linewidth]{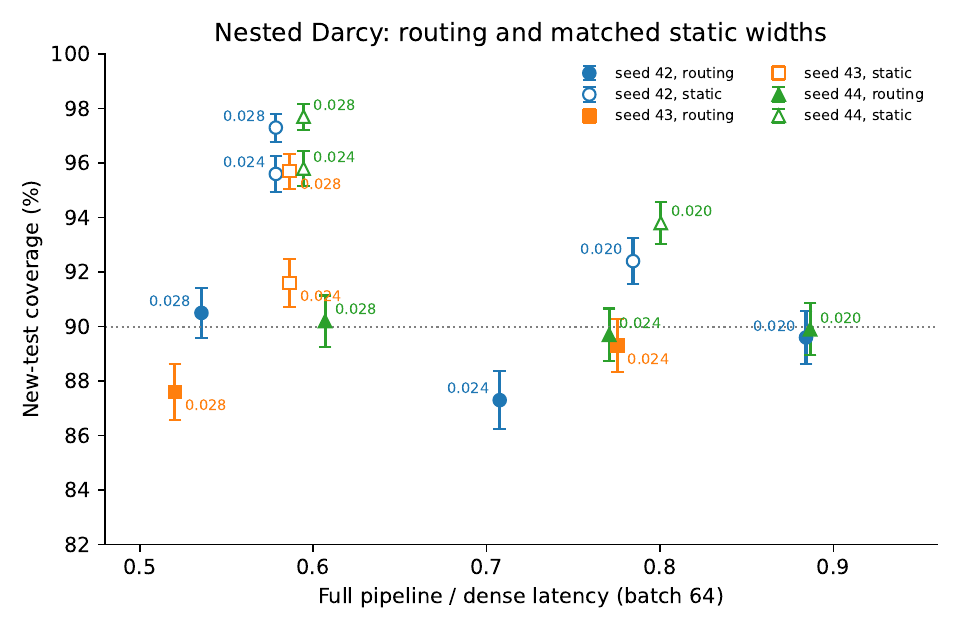}
\caption{New-test coverage against full-pipeline latency measured on original
canonical batches at batch 64; point labels are $\tau$. Error bars are
$\pm1$ binomial standard error at $n=1000$ (about 0.95 percentage points
near 90\%). Open markers denote matched static policies; filled markers
denote routing. Coverage and timing use separate samples. Static points
are shifted horizontally by seed for legibility; all static width-16
points have ratio 0.586 and all static width-24 points 0.792.}
\label{fig:width-cost-coverage}
\end{figure}

\subsubsection{Complete runtime under matched nominal guarantees}
\label{sec:width-ceiling-cost}
At batch size 64, the nested selector's full pipeline is faster than its
matched static model at two of three seeds for $\tau=0.028$ and is
3.6\% slower at seed 44 (0.607 versus 0.586 times dense). At $\tau=0.020$
and $0.024$, matched static inference is faster at every feasible nested
cell, and on the timing batches dispatch-only measurements also lose at
every $\tau=0.024$ cell, even before scoring is included. Those batches
under-represent width-8 routing (Section~\ref{sec:width-selector});
Appendix~\ref{app:dispatch} shows that reweighting to the new-test
composition leaves every $\tau=0.024$ complete pipeline slower than its
static comparator, and estimates the effect of an unmeasured GPU-scoring
pipeline. For independently trained models on the
same A6000 node, the selector is clearly faster at three of nine
independent-family settings, static at four, and the remaining two differ
by at most 1.5 percentage points of dense latency. Two of the four
apparent static wins have infinite routing thresholds; those constant
width-8 routes can bypass the router and tie static width 8. Counting
these as ties leaves two static wins, three routing wins, two near ties,
and two exact ties. Thus neither model family shows a broad runtime
advantage for per-input routing; the result depends on the budget and
which fixed model is selected.
The cost-order score $h(w)=-w$ is one valid static ranking, but
independently trained errors need not be nested. Appendix~\ref{app:validation-ranked-static}
reports a second ordering fixed by validation error quantiles. At seed 44
and $\tau=0.024$, it selects width 8; coverage is 93.6\% versus 93.8\%
for routing, and static latency is 0.383 versus 0.467 times dense.
Across these nine independent settings, this static policy is faster at
six, routing at two, and one is within 1.5 percentage points of dense
latency.
\FloatBarrier

\section{Discussion}
\label{sec:discussion-static}

The experiments separate calibrated error control from measured speed.
Mode truncation reduces spectral work while leaving full-grid transforms
and pointwise channel mixing intact; in these measurements, that changed
latency little. Narrowing channels reduces mixing and transform costs,
but reference-based selection pays for the dense prediction. Input-only
routing avoids that pass, while its measured gain depends on the input
score, action set, and dispatch implementation.

Three factors limit the Darcy router, and our experiments do not separate
them: a single hand-crafted statistic, $\mathrm{TV}(x)$; a binary action
set, whereas static policies can choose widths 16 or 24 at tighter
budgets; and the 19.3\% weight-sharing penalty of the nested width-8
slice. The direct-score rule accepts any frozen score $h(x,w)$, so a
lightweight learned score over all widths, calibrated on independent
data, is a natural next experiment.

The coverage results illustrate the scope of split-conformal validity.
The guarantee averages over calibration and deployment draws, and with
624 calibration examples the coverage of one fitted policy can plausibly
fall a few points below target. At the tightest spectral budget the
direct-score shortfall exceeds that variability, suggesting a small
tail mismatch between the regenerated Darcy set and the canonical pool
(Appendix~\ref{app:spectral-resplit}). Practitioners who need control for the
calibration set at hand should use larger calibration sets or
high-probability risk control, which targets a different statistical
guarantee.

Our results concern the tested FNO architectures, resolutions, batch
sizes, and GPU. Higher-resolution two-dimensional problems, three-
dimensional or time-dependent PDEs, and fused input-only routers would
test whether spectral contraction becomes a larger share of execution
and whether finer-grained routing repays its scoring and dispatch costs.

Calibration also does not protect against changes in the deployment
distribution. Coverage collapses in both severe stress tests of
Appendix~\ref{app:shift}: Burgers at viscosity $\nu=0.001$ or $\nu=0.1$
instead of the training value $0.01$ (0\% coverage), and Darcy
coefficients spanning 1--1000 instead of $\{3,12\}$ (0\% for the seed-42
TV selector at every budget). The triangle bound still holds, but the
full model's own error lies far above its calibration quantile, so
selecting more support or width cannot restore accuracy. A range check on
input total variation flags every shifted Darcy input; this is an OOD
alarm, not an accuracy guarantee. Existing remedies
for non-exchangeable data do not directly apply to the shifts tested
here. Weighted conformal prediction \citep{tibshirani2019covariateshift}
requires a likelihood ratio between test and calibration inputs, but
the shifted Darcy coefficients lie outside the calibration support, and
the Burgers viscosity shift changes the solution operator rather than
the input distribution. Adaptive conformal inference
\citep{gibbs2021aci} requires labelled feedback, which here means a
numerical PDE solve per checked input. Combining the selection rules
with such methods under milder, overlapping shifts is left to future
work.

\Needspace{16\baselineskip}
\section{Conclusion}

We introduced simultaneous reference envelopes and a direct-score rule
for selecting computation in neural operators, including a static policy
calibrated with the same direct-score construction. The experiments show
that selected support alone does not predict latency: mode truncation
was nearly latency-neutral in the tested sweeps, while width reduction
lowered Darcy forward cost. Input-only routing sometimes improved on a
matched static policy, but its binary action set and execution overhead
limited that advantage. Future comparisons should report achieved
coverage and complete runtime against a family-matched static policy
under the same calibration protocol.

\section*{Data and code availability}
The code repository (\coderepo) contains experiment code, archived
per-instance errors and timing summaries, figure sources, and reproduction
instructions. It also includes a note describing the material shared with
the compression study \citep{ibrahim2026selection}. A verification script checks the
supported manuscript quantities against the archived evidence without a
GPU, raw PDE datasets, or checkpoints. The repository documents the
independent Darcy generator, fixed calibration pool, and development
choices, including the excluded pilot with incorrect coefficient levels.

The trained model checkpoints for all seeds (dense, nested-width, and
independently trained width models) and the separately generated
1,000-example Darcy test set will be released in a public archive upon
publication, so that the reported errors can be
reproduced and the timing measurements rerun without retraining. Raw datasets are not
included in the repository. The supplied commands
cover data preparation, training, calibration, and evaluation; exact
retraining is not guaranteed across hardware and library versions. The
archived errors reproduce the reported statistical analyses, while new
training and timing runs constitute replication. The timing sample and
new coverage sample are explicitly distinguished.

\bibliographystyle{tmlr}
\bibliography{references}

\clearpage
\appendix

\section{Calibration and implementation details}

\subsection{Numerical and infeasibility safeguards}
\label{app:safeguards}
When $r=\lceil(1-\delta)(n+1)\rceil>n$, the conformal upper bound is
$+\infty$, not the largest observed score. Using the maximum instead
would be wrong: with a single continuous IID calibration score, the
maximum covers a new score with probability only one half, not 90\%.
Lower bounds take the corresponding $-\infty$ convention by negation.
This case does not arise at the sample sizes and risk levels we report.
Regression tests cover a rank equal to $n$, a rank above $n$, infinite
scores, and invalid inputs.

The direct-score rule treats a candidate as eligible only if
$h(x,K)<\hat\lambda$, which excludes ties, because the conformal event
is $s(x)\ge\hat\lambda$. A threshold of $-\infty$ means calibration is
infeasible, and the rule then abstains rather than quietly returning
the dense prediction as certified. The fresh-split, input-only, group,
and width-contamination evaluations all use the same convention.

The triangle rule allows equality but adds no positive slack (such as
$10^{-9}$) to $\tau-\hat q_\delta$, so at an ambiguous floating-point boundary
the smaller action is rejected. An earlier version of our code added
this slack, which enlarges the tolerance. Re-evaluating the
spectral tables and the eight feasible fresh-pool width points under the
current comparisons leaves their reported outcomes unchanged, and
deterministic tests cover tied scores and infeasible calibration.

\subsection{Why scoring only the smallest accurate candidate is unsafe}
\label{app:robust-score-unsound}
Suppose each calibration example is scored by its disagreement at the
smallest accurate support, and that score is used as a one-sided
cutoff. An inaccurate candidate with even smaller disagreement can
then pass the cutoff. Nesting the spectral multipliers does not make
the nonlinear predictions, their errors, or their disagreements
monotone in support.

In the recorded grids, disagreement is non-monotone on 8\% of Burgers
examples and on 11.5--60\% of Darcy examples, depending on the seed. At
Darcy $\tau=0.05$, this naive rule calibrates to $+\infty$ on two
seeds, always selects the cheapest support, and has 0\% evaluation
coverage. The direct score in Eq.~\eqref{eq:robust-score} avoids this
by taking the minimum score over every inaccurate candidate; with the
strict cutoff, all of them are excluded on the conformal event.

\section{Supplementary spectral analyses}

\subsection{Re-split diagnostic for held-out spectral coverage}
\label{app:spectral-resplit}
For each Darcy cell, we pooled the calibration ($n=624$) and evaluation
($n=1000$) direct scores and drew 2000 random splits of the same sizes,
which are exchangeable by construction. The rate of the conformal event
$s(x)\ge\hat\lambda$ on the evaluation part then has mean 90.0--90.1\% and
standard deviation about 1.5 points, matching the calibration-conditional
analysis of Section~\ref{sec:width-selector}. At $\tau=0.05$ the observed
event rates of 85.4\%, 88.0\%, and 86.7\% fall at the 0.1\%, 10.3\%, and
2.2\% points of this distribution for seeds 42, 43, and 44; the only
other value below 5\% is seed 44 at $\tau=0.18$ (3.6\%). Marginal
two-sample KS tests of the dense error and of the candidate errors and
disagreements at every support show no detectable shift between the two
sources (all $p\ge0.19$). With 12
correlated cells, and seeds sharing the same examples, these values are
not independent tests. They suggest a small difference in the joint
error tail between the canonical pool and the regenerated set that
matters most at the tightest budget; we do not identify its cause. The
same re-split check on the width selector cells of
Table~\ref{tab:matched-static} gives no comparable outlier (smallest
values 4\% and 6\% across 17 cells). The triangle rule, whose event
depends only on the dense error, covers 99.5--100\% on the same data.

\subsection{Preliminary relative-error oracle diagnostic}
\label{sec:oracle}
Before building an estimator, we checked whether the minimum accurate
support varies across examples at all. This diagnostic uses the \emph{relative} error
$r_K(x)=\lVert u_K-u^\star\rVert_2/\lVert u^\star\rVert_2$, not the
absolute error used for calibration. It selects
$K_{\mathrm{rel}}^\star(x)=\min\{K:r_K(x)\le\rho(x)\}$, where either
$\rho=(1+\alpha)r_{\Kmax}$ for a relative slack $\alpha$, or
$\rho=\epsilon$ is a fixed ceiling on relative error. If no candidate
passes, the diagnostic assigns $\Kmax$, without claiming that it is
accurate. Variability under this relative-error diagnostic does not by
itself imply variability under the fixed absolute budgets used by the
calibrated selector. On the Burgers validation split (200 instances, seed 42, no
retraining), across five tolerance settings, 3--57\% of instances
needed the full support, and mean $K_{\mathrm{rel}}^\star$ ranged from
10.53 to 14.91 out of 16 (Table~\ref{tab:oracle}). The diagnostic
treated variability as negligible if at least 95\% of instances needed
the full support; none of the five settings reached this.

\begin{table}[!htbp]
\centering
\caption{Relative-error oracle support, Burgers, seed 42, validation
($n=200$). The fixed ceilings are thresholds on relative error, not the
absolute budgets used for calibration.}
\label{tab:oracle}
\begin{tabular}{lrrrr}
\toprule
Tolerance & Mean $K_{\mathrm{rel}}^\star$ & Median $K_{\mathrm{rel}}^\star$ & At $K=16$ & Support fraction \\
\midrule
relative-slack 0.10 & 14.91 & 16 & 57.0\% & 93.2\% \\
relative-slack 0.25 & 13.53 & 14 & 17.5\% & 84.6\% \\
relative-slack 0.50 & 12.33 & 12 & 3.0\% & 77.1\% \\
fixed $\epsilon{=}0.005$ & 10.53 & 10 & 9.5\% & 65.8\% \\
fixed $\epsilon{=}0.003$ & 12.61 & 12 & 25.0\% & 78.8\% \\
\bottomrule
\end{tabular}
\end{table}

\FloatBarrier
\subsection{Complexity analysis on the non-saturated subset}
\label{app:range-restriction}
This analysis uses the same validation-split selections as
Section~\ref{sec:complexity}. At tight Darcy tolerances, the selected $K$ is at or near $\Kmax$ for
68--92\% of instances, which leaves little variance to correlate with.
We therefore also compute correlations on the non-saturated instances.
Because this subset is selected on the outcome, these correlations are
only a diagnostic, not an estimate of the population correlation.

At $\tau=0.18$, only 1--9.5\% of instances saturate, and the subset
correlations range from $-0.351$ to $-0.301$. We report these as
descriptive correlations only: conditioning on the selected support
truncates the outcome and invalidates the usual Fisher-$z$ interval
interpretation. At $\tau=0.08$, the subset correlations are positive for
all seeds, but the samples are small. The relationship at tight
tolerances remains unresolved.

\section{Distribution-shift stress tests}
\label{app:shift}

\subsection{Burgers viscosity shift}
\label{sec:shift}
\label{sec:severe-shift}
We keep the Burgers model, normalization, and in-distribution quantile
fixed and evaluate at viscosities $\nu=0.001$ and $\nu=0.1$, compared
with $\nu=0.01$ in training. Coverage drops to zero under both shifts
(Table~\ref{tab:ood}). The triangle bound still holds, but the
reference error grows far beyond its calibration quantile.
Each shifted evaluation uses the 200-example test cache for the changed
viscosity, with the same Burgers initial-condition construction as the
benchmark and in-distribution normalization statistics. We evaluate the
stored targets and do not rerun or independently audit the solver for
these severe shifts.
The policy selects mean supports of 7.31 and 11.90 modes, respectively;
thus the shifted inputs do not all escalate to full support.

\begin{table}[H]
\centering
\caption{Certificate under severe out-of-distribution viscosity shift ($\tau = 0.03$, $\delta = 0.10$; 200 examples per shifted dataset). The stress test keeps
the seed-42 quantile calibrated on the validation split; the
in-distribution row evaluates all 200 validation examples.}
\label{tab:ood}
\begin{tabular}{lrrrrr}
\toprule
Dataset & $e_{\max}$ mean & vs.\ in-distribution & Mean $\hat K$ & Held-out coverage & Target \\
\midrule
In-distribution, $\nu{=}0.01$ & 0.0042 & $1\times$ & 7.31 & 100.0\% & 90\% \\
Out-of-distribution, $\nu{=}0.001$ & 1.074 & $\sim\!254\times$ & 7.31 & 0.0\% & 90\% \\
Out-of-distribution, $\nu{=}0.1$ & 4.865 & $\sim\!1151\times$ & 11.90 & 0.0\% & 90\% \\
\bottomrule
\end{tabular}
\end{table}

We see this failure only because the evaluation data are labelled. The
deployment rule does not detect it, since a candidate and the reference
can agree while sharing a large error. For a mix of clean and shifted
inputs, coverage follows $C(p)=(1-p)C_{\mathrm{ID}}+pC_{\mathrm{OOD}}$,
which here is approximately $1-p$ (Appendix~\ref{app:shift-mixtures}).
Appendix~\ref{sec:width-robustness} repeats the test for width routing.

\subsection{Mixtures of in-distribution and shifted inputs}
\label{app:shift-mixtures}
\begin{figure}[!htbp]
\centering
\includegraphics[width=0.62\linewidth]{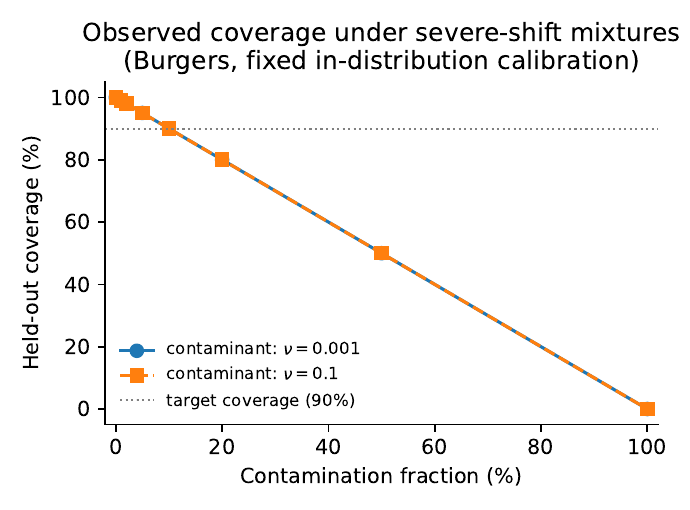}
\caption{Held-out coverage against contamination fraction, Burgers,
both severe-shift sources. Coverage is 90\% at 10\% contamination and
80\% at the next sampled level, 20\%.}
\label{fig:contamination}
\end{figure}

The Burgers contamination sweep mixes in-distribution examples with
examples from the two severe viscosity shifts of
Appendix~\ref{sec:shift}, keeping the policy and its in-distribution
quantile fixed. The clean examples pass and the shifted ones fail, so
coverage follows $C(p)=(1-p)C_{\mathrm{ID}}+pC_{\mathrm{OOD}}\approx1-p$
(Figure~\ref{fig:contamination}). This describes the curve; it is not a
guaranteed tolerance to contamination.

Because both shifts are severe, a small contamination fraction is not
the same as a mild change in viscosity. We did not test mild physical
shifts, or spectral selection on shifted Darcy data.
\FloatBarrier

\subsection{Darcy coefficient shift and width routing}
\label{sec:width-robustness}
We keep the models and in-distribution thresholds fixed and draw Darcy
coefficients from a multiscale generator with values 1--1000, compared
with 3--12 in the canonical data. In the recorded experiments, the
reference-based rule assigns every shifted input to full width, yet
coverage collapses. The seed-42 TV selector likewise sends every shifted
input to width 32 and has 0\% coverage at every reported budget.

The shifted inputs have total variation 8.6 on average (range
4.8--16.2), against 0.09--0.78 in the fresh calibration pool. A range check
at its maximum, 0.78, flags every shifted example with no model forward
pass; this is an OOD alarm, not an accuracy guarantee.
Both selection signals also send these inputs to full width, but the full
model is itself wrong, so escalating cannot restore accuracy.

\section{Supplementary width analyses}
The model-accuracy, reference-selection, and static-width-8 diagnostics
below use the original 200-example test split inspected during
development. The independently trained selector comparison uses the new
test set, as stated in its caption.

\subsection{Descriptive static-width screen}
\label{app:static-screen}
The following calibration-only screen reports the cheapest width whose
individual quantile clears the budget. Its selection is descriptive;
the main comparison uses the matched direct-score baseline of
Section~\ref{sec:width-ceiling-bonferroni}. For three independent-model
cells (seed 42 at $\tau=0.020$ and seed 44 at $\tau=0.020$ and $0.024$),
the screen picks a narrower width than Table~\ref{tab:matched-static},
because the direct-score rule counts a calibration example in favor of a
width only if every wider width is also within budget on that example.

\begin{table}[H]
\centering
\caption{Smallest width passing the fresh-pool quantile screen
($n=624$, $\delta=0.10$), by model family and training seed.
A dash means no width passes.}
\label{tab:width-ceiling}
\begin{tabular}{lrrrrrr}
\toprule
& \multicolumn{3}{c}{Nested checkpoint} & \multicolumn{3}{c}{Independent models}\\
\cmidrule(lr){2-4}\cmidrule(lr){5-7}
$\tau$ & 42 & 43 & 44 & 42 & 43 & 44\\
\midrule
0.020 & 24 & -- & 24 & 16 & 24 & 24\\
0.024 & 16 & 16 & 16 & 16 & 16 & 8\\
0.028 & 16 & 16 & 16 & 8 & 16 & 8\\
\bottomrule
\end{tabular}
\end{table}

\subsection{Nested training and the accuracy cost of sharing weights}
\label{app:width-training}
The nested Darcy FNO has no normalization layers, so switchable batch
normalization is unnecessary. At every training step, losses from all
four widths are summed before a single backward pass. Validation error
over the whole set decreases with width at the logged epochs, although
errors on individual examples need not follow that ordering.

\begin{figure}[!htbp]
\centering
\includegraphics[width=0.68\linewidth]{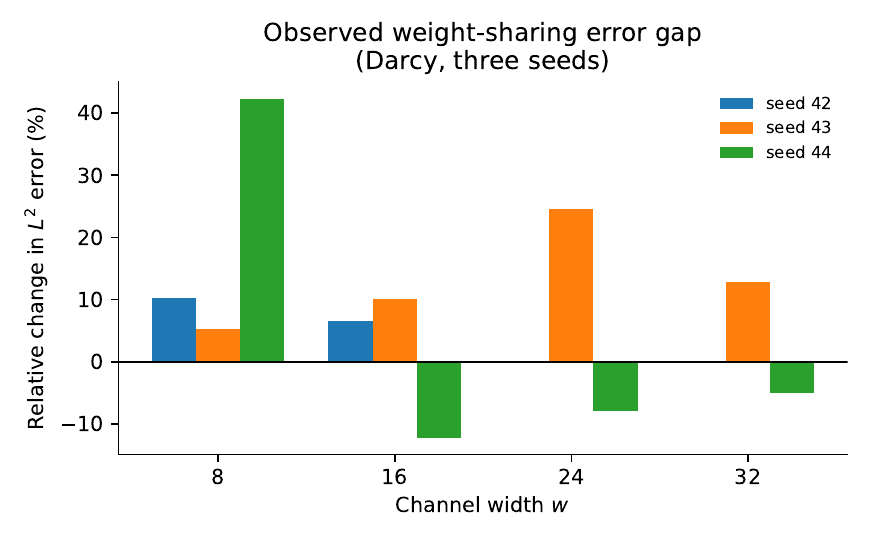}
\caption{Relative error gap between nested and independently trained
models of the same width, Darcy, by seed.}
\label{fig:width-shared-cost}
\end{figure}

Full-width test errors stay within our 15\% relative acceptance
threshold against dense training. Table~\ref{tab:width-test-error}
gives the errors at every width. The dedicated width-8, 16, and 24
models use the same hyperparameters; the independent width-32 entries
are the existing dense baselines. Figure~\ref{fig:width-shared-cost} compares them with the
nested slices of the same width.

\begin{table}[!htbp]
\centering
\caption{Original-test relative $L^2$ error by seed for nested slices
and independently trained Darcy models. Independent narrow-width values
are included to show the weight-sharing gap directly.}
\label{tab:width-test-error}
\scriptsize
\resizebox{\linewidth}{!}{%
\begin{tabular}{crrrrrrrr}
\toprule
& \multicolumn{4}{c}{Nested checkpoint} & \multicolumn{4}{c}{Independently trained}\\
\cmidrule(lr){2-5}\cmidrule(lr){6-9}
Seed & $w=8$ & $w=16$ & $w=24$ & $w=32$ & $w=8$ & $w=16$ & $w=24$ & $w=32$\\
\midrule
42 & 0.021301 & 0.014151 & 0.012612 & 0.012098 & 0.019331 & 0.013280 & 0.012634 & 0.012107\\
43 & 0.021760 & 0.016278 & 0.014320 & 0.013673 & 0.020660 & 0.014789 & 0.011494 & 0.012116\\
44 & 0.022569 & 0.013855 & 0.012242 & 0.011429 & 0.015863 & 0.015779 & 0.013298 & 0.012040\\
\bottomrule
\end{tabular}}
\normalsize
\end{table}

The width-8 gap is positive for all three seeds and averages 19.3\%;
at larger widths the sign varies. Table~\ref{tab:width-test-error} reports both model families. The penalty at the smallest width is
therefore consistent, although three seeds say little about its
variability.

On the 200-example grids, disagreement never increases with width,
up to a numerical tolerance of $10^{-9}$, whereas the true errors are
non-monotone on 21--32\% of examples. Sharing prefix weights guarantees
neither ordering, and the direct-score rule relies on neither.
\FloatBarrier

\subsection{Reference-based selection: error scale and action variability}
\label{app:width-reference}
Table~\ref{tab:tau-meaning} expresses the earlier reference-based
results relative to the dense model's mean error. At $\tau=0.028$, the
mean error of the selected widths is 1.5--1.8 times the mean dense
error, with 83--93\% of inputs at width 8. These ratios help interpret the absolute budgets,
although a fixed absolute tolerance does not correspond to a fixed
relative error.

\begin{table}[!htbp]
\centering
\caption{$\eta$-based rule: mean error of the chosen width relative to
the mean dense error, and fraction routed to width 8. Calibrated on the
earlier $n=200$ checkpoint-selection pool and evaluated on the test
split.}
\label{tab:tau-meaning}
\begin{tabular}{lrrr}
\toprule
Seed & $\tau$ & Mean chosen error / dense error & Frac.\ at width 8 \\
\midrule
42 & 0.020 & $1.08\times$      & $0.0\%$  \\
42 & 0.024 & $1.46\times$      & $64.0\%$ \\
42 & 0.028 & $1.66\times$      & $93.0\%$ \\
43 & 0.020 & infeasible        & --       \\
43 & 0.024 & $1.34\times$      & $47.5\%$ \\
43 & 0.028 & $1.48\times$      & $83.0\%$ \\
44 & 0.020 & $1.29\times$      & $17.5\%$ \\
44 & 0.024 & $1.68\times$      & $77.0\%$ \\
44 & 0.028 & $1.81\times$      & $91.0\%$ \\
\bottomrule
\end{tabular}
\end{table}

For comparison, consider the label-informed policy
$w^\star(x)=\min\{w:e_w(x)\le\tau\}$, which assigns width 32 when no
width is accurate. Its cost proxy is about $0.51\times$ dense at
$\tau=0.020$, $0.45$--$0.46\times$ at 0.024, and $0.43\times$ at 0.028.
This reflects how much the accurate width varies across inputs. It is
not a lower bound on cost at matched risk, since a policy allowed a
fraction $\delta$ of errors can pick cheaper, inaccurate widths.

Recalibrating this rule on the fresh pool gives some test coverages
below target; Appendix~\ref{app:width-selector-robustness} gives the
sampling diagnostics.
\FloatBarrier

\subsection{Static nested width 8}
\label{app:static-w8}
Table~\ref{tab:static-w8} compares the nested width-8 slice used on its
own with the nested binary selector. Width 8 alone runs at
$0.38$--$0.40\times$ dense latency in isolated batched measurements,
and its upper error quantile on the fresh calibration pool is about
0.029--0.030. The selector's coverage advantage is largest at the
tighter budgets.

\begin{table}[!htbp]
\centering
\caption{Original-test coverage of the nested width-8 slice alone and of the
nested binary selector. Ranges are across seeds; the selector columns
omit the infeasible seed-43 row at $\tau=0.020$. Cost is the
selector's isolated-width proxy, not complete runtime.}
\label{tab:static-w8}
\begin{tabular}{lrrr}
\toprule
$\tau$ & Static $w=8$ coverage & Selector coverage & Selector cost proxy\\
\midrule
0.020 & $57.5$--$64.0\%$ & $88.0$--$89.0\%$ & $0.76$--$0.77\times$\\
0.024 & $78.0$--$80.5\%$ & $89.5$--$90.5\%$ & $0.55$--$0.62\times$\\
0.028 & $85.5$--$88.5\%$ & $87.0$--$91.5\%$ & $0.42$--$0.47\times$\\
\bottomrule
\end{tabular}
\end{table}

This slice fails the quantile screen at all three budgets, but that
says nothing about static models in general. Static width 16 and the
independently trained narrow models in Table~\ref{tab:width-ceiling}
are stronger static comparators; Section~\ref{sec:width-ceiling-bonferroni}
reports their matched direct-score calibration.
\FloatBarrier

\subsection{Validation-ranked static score for independent widths}
\label{app:validation-ranked-static}
The cost-order score $h(w)=-w$ is simple, but it can be conservative for
independently trained widths whose errors are not nested. As a sensitivity
analysis, we set each non-maximal action's input-independent score to its
90th error percentile on the validation split, frozen before calibration.
We then apply Eq.~\eqref{eq:robust-score} to the same $n=624$ calibration
pool with $\delta=0.10$ and the same full-width $-\infty$ convention.
The resulting fixed action has the same marginal error-event guarantee;
the fresh 1,000-example test set is used only for evaluation.

\begin{table}[!htbp]
\centering
\caption{Independent-family static policies ranked by validation error
quantiles, then direct-score calibrated on the fresh $n=624$ pool. Coverage
uses the new $n=1000$ Darcy test set. Latencies are A6000 batch-64 ratios;
static timing is a single independent-model forward pass, while selector
timing includes scoring and dispatch.}
\label{tab:validation-ranked-static}
\small
\begin{tabular}{rrrrrr}
\toprule
Seed & $\tau$ & Static width & Static coverage & Static / dense & Selector / dense\\
\midrule
42 & 0.020 & 16 & 93.5\% & 0.586$\times$ & 0.786$\times$\\
42 & 0.024 & 16 & 96.6\% & 0.586$\times$ & 0.589$\times$\\
42 & 0.028 & 8  & 91.5\% & 0.384$\times$ & 0.466$\times$\\
43 & 0.020 & 24 & 95.2\% & 0.792$\times$ & 0.834$\times$\\
43 & 0.024 & 16 & 93.9\% & 0.586$\times$ & 0.686$\times$\\
43 & 0.028 & 16 & 96.4\% & 0.586$\times$ & 0.479$\times$\\
44 & 0.020 & 32 & 94.0\% & 1.000$\times$ & 0.774$\times$\\
44 & 0.024 & 8  & 93.6\% & 0.383$\times$ & 0.467$\times$\\
44 & 0.028 & 8  & 96.6\% & 0.383$\times$ & 0.467$\times$\\
\bottomrule
\end{tabular}
\end{table}

Under this ranking, static coverage is higher in six cells, equal in the
two cells where both policies use width 8, and lower by 0.2 percentage
points in one. The static policy is
faster at four settings, routing at two, one timing difference is
within 1.5 percentage points of dense latency, and two infinite-threshold
cells tie after bypassing the router. At seed 44 and
$\tau=0.024$, this ordering removes the routing advantage reported by
the cost-ordered baseline. The comparison shows that runtime conclusions
depend on the fixed static ranking, even though both rankings are
calibrated independently of the test set.
\FloatBarrier

\subsection{Calibration over the deployed binary action set}
\label{app:binary-action-audit}
The main binary selector's score includes errors at widths 16 and 24,
even though deployment uses only widths 8 and 32. We recomputed the
locked calibration and evaluation using the two-action score
$s_{\mathrm{bin}}$ defined in Section~\ref{sec:width-selector}, from
the archived $n=624$ calibration errors and the new $n=1000$ test errors.
For nested weights, no calibration example had both widths 8 and 32
within budget while an unused intermediate width failed, so the threshold,
routing fractions, and coverage are unchanged in all nine seed--budget
cells. For independently trained models, such examples occurred in at
most 6.1\% of the calibration pool. The threshold changed in five of
nine cells; on the new test set the resulting width-8 routing fraction
changed by at most 8.9 percentage points, while coverage changed by at
most 0.5 points. The largest routing change was seed 44 at
$\tau=0.020$; coverage moved from 92.8\% to 92.3\%. Thus, unused-width
penalties can affect calibration for independent models, although this
ablation does not remove the binary router's restricted action set or
include new latency measurements. Per-cell recalibrated results are
archived in the code repository.

\subsection{Independent-model routing}
\label{app:two-model-selector}
Independent-model routing results and width-8 routing fractions are included
in the master comparison, Table~\ref{tab:matched-static}. At
$\tau=0.028$ for seeds 42 and 44, the fitted threshold is infinite and
all inputs use width 8. Those constant-route policies can bypass scoring
and dispatch; the integrated timings in the table retain the generic
pipeline and therefore overstate their deployed cost.

\section{Timing details}

\subsection{Scoring, dispatch, and timing boundaries}
\label{sec:tv-cost}
The dispatch-only measurements take the total variation and its
threshold comparison as precomputed. To include this work, we also
timed scoring on its own and the complete pipeline: CPU NumPy scoring,
threshold transfer, GPU sub-batch dispatch, and recombination, with
the same timing protocol as the main experiments, for all 24 feasible
combinations of seed, budget, and batch size.

The complete pipeline takes $0.520$--$0.948\times$ dense latency.
CPU-loop scoring costs $9.98$--$11.25\%$ of dense latency ($1.1$--$1.2$ms
at batch 20 and $7.0$--$7.3$ms at batch 128). Timing starts with the
raw input on the CPU and the normalized batch on the GPU, so data
loading and normalization are excluded for both policies.

We also timed vectorized CPU scoring ($5.72$--$9.73\%$ of dense) and
GPU scoring on its own ($0.29$--$1.21\%$, $0.13$--$0.19$ms). Adding the
GPU scoring time to the measured dispatch time gives
$0.416$--$0.856\times$ dense. This is an \emph{additive estimate}, not a
measurement of a combined implementation: it assumes the raw input is
already on the GPU, and it does not account for extra synchronization
or check that GPU scoring gives the same routing decisions.

The GPU estimate indicates what a combined GPU implementation might
gain; we have not measured one.

\subsection{Dispatch measurements and batch composition}
\label{app:dispatch}
\begin{figure}[!htbp]
\centering
\includegraphics[width=0.95\linewidth]{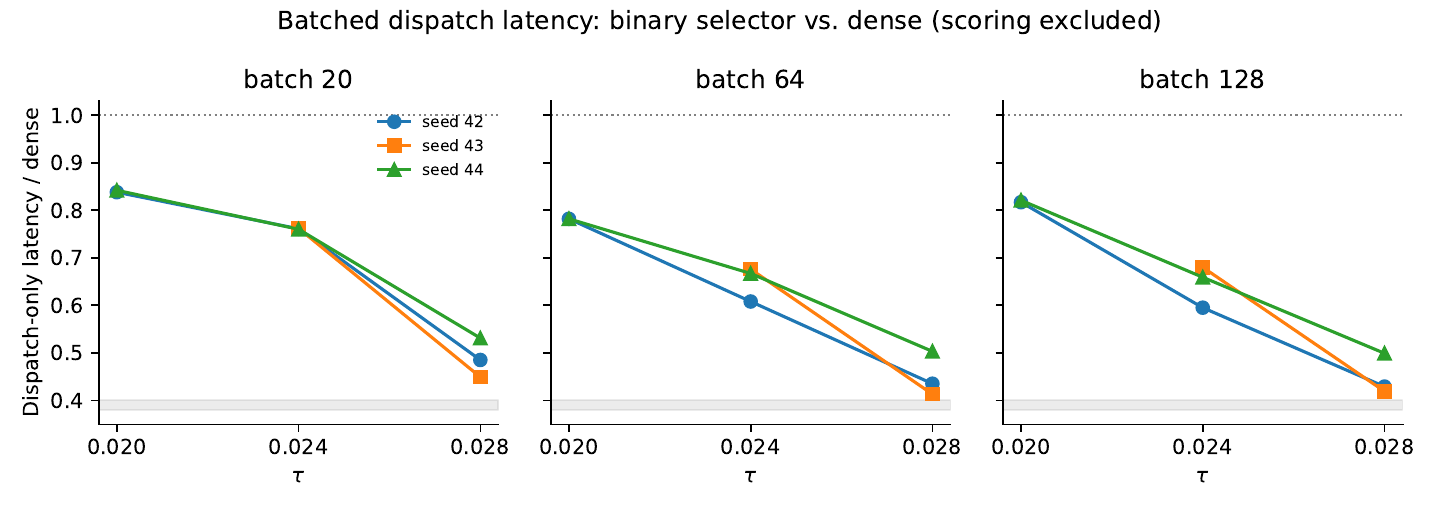}
\caption{Dispatch-only latency ratio of the binary selector, all three
seeds, by batch size, excluding scoring. The shaded band is the latency
of nested width 8 alone, which is not accuracy-matched.}
\label{fig:e2e-latency}
\end{figure}

The dispatch-only timing covers bucketing, the two sub-batch forward
passes, and recombination, with scores precomputed.
Figure~\ref{fig:e2e-latency} shows the ratio for each batch. The
composition-matched proxy weights the isolated-width latencies by the
width-8 fraction of each timed batch, rather than of the whole test
set. The remaining dispatch overhead is larger at batch 20 than at
batches 64 and 128, consistent with better amortization at larger
batch sizes.

At batch 64 and $\tau=0.024$, the timing batches send 7--10 points
fewer inputs to width 8 than the new test set
(Section~\ref{sec:width-selector}). Reweighting with the
composition-matched proxy to the new-test fractions would put seed 42's
dispatch-only time at roughly $0.56\times$ dense, below static width 16,
but adding the measured CPU scoring still leaves every $\tau=0.024$
complete pipeline slower than its static comparator. Conversely, with
the additive GPU-scoring estimate of Appendix~\ref{sec:tv-cost}, seed 44
at $\tau=0.028$ would be about $0.51\times$ dense and faster than static
width 16; that combined implementation was not measured.

\subsection{Per-batch latency of the binary width selector}
\label{app:e2e-table}
Table~\ref{tab:e2e-latency} gives the per-batch numbers behind
Figure~\ref{fig:e2e-latency}, including the composition-matched proxy
from Appendix~\ref{app:dispatch}.

\begin{table}[!htbp]
\centering
\caption{Dispatch-only batched wall-clock latency of the binary
selector and the dense model, all three seeds, trained weights,
single GPU,
synchronized, on the original canonical test batches, with the same thresholds from the $n{=}624$ calibration pool. Scoring is
precomputed and excluded; complete-pipeline ranges are given in
Section~\ref{sec:tv-cost}. The last column
gives the composition-matched proxy and the measured ratio. Standard
deviations over repeats are at most 0.5ms in every cell and are
omitted; the per-seed timing records in the code repository include them.}
\label{tab:e2e-latency}
\scriptsize
\begin{tabular}{rrrrrrr}
\toprule
Seed & $\tau$ & Batch & Dense (ms) & Selector (ms) & Local frac.\ @ w8 & Proxy ratio / Measured ratio \\
\midrule
42 & 0.020 & 20 & 11.08 & 9.29 & 45.0\% & 0.730$\times$ / 0.838$\times$ \\
42 & 0.020 & 64 & 34.59 & 27.05 & 37.5\% & 0.768$\times$ / 0.782$\times$ \\
42 & 0.020 & 128 & 64.72 & 52.86 & 39.1\% & 0.761$\times$ / 0.817$\times$ \\
42 & 0.024 & 20 & 11.10 & 8.44 & 55.0\% & 0.670$\times$ / 0.761$\times$ \\
42 & 0.024 & 64 & 34.70 & 21.10 & 65.6\% & 0.595$\times$ / 0.608$\times$ \\
42 & 0.024 & 128 & 64.77 & 38.52 & 72.7\% & 0.555$\times$ / 0.595$\times$ \\
42 & 0.028 & 20 & 11.13 & 5.40 & 90.0\% & 0.459$\times$ / 0.485$\times$ \\
42 & 0.028 & 64 & 34.79 & 15.14 & 93.8\% & 0.421$\times$ / 0.435$\times$ \\
42 & 0.028 & 128 & 64.82 & 27.80 & 95.3\% & 0.417$\times$ / 0.429$\times$ \\
43 & 0.024 & 20 & 11.15 & 8.49 & 55.0\% & 0.670$\times$ / 0.762$\times$ \\
43 & 0.024 & 64 & 34.85 & 23.55 & 54.7\% & 0.662$\times$ / 0.676$\times$ \\
43 & 0.024 & 128 & 64.84 & 44.13 & 60.2\% & 0.632$\times$ / 0.680$\times$ \\
43 & 0.028 & 20 & 11.17 & 5.02 & 95.0\% & 0.429$\times$ / 0.449$\times$ \\
43 & 0.028 & 64 & 34.91 & 14.42 & 96.9\% & 0.401$\times$ / 0.413$\times$ \\
43 & 0.028 & 128 & 64.91 & 27.22 & 96.9\% & 0.407$\times$ / 0.419$\times$ \\
44 & 0.020 & 20 & 11.17 & 9.40 & 45.0\% & 0.730$\times$ / 0.842$\times$ \\
44 & 0.020 & 64 & 34.91 & 27.31 & 37.5\% & 0.768$\times$ / 0.782$\times$ \\
44 & 0.020 & 128 & 64.91 & 53.28 & 39.1\% & 0.761$\times$ / 0.821$\times$ \\
44 & 0.024 & 20 & 11.20 & 8.51 & 55.0\% & 0.670$\times$ / 0.760$\times$ \\
44 & 0.024 & 64 & 34.96 & 23.30 & 56.3\% & 0.652$\times$ / 0.667$\times$ \\
44 & 0.024 & 128 & 64.95 & 42.80 & 63.3\% & 0.613$\times$ / 0.659$\times$ \\
44 & 0.028 & 20 & 11.21 & 5.96 & 85.0\% & 0.489$\times$ / 0.531$\times$ \\
44 & 0.028 & 64 & 34.98 & 17.60 & 82.8\% & 0.488$\times$ / 0.503$\times$ \\
44 & 0.028 & 128 & 64.96 & 32.42 & 85.2\% & 0.479$\times$ / 0.499$\times$ \\
\bottomrule
\end{tabular}
\normalsize
\end{table}

\section{Calibration sensitivity and distribution diagnostics}
\label{app:width-selector-robustness}

\paragraph{Main new test ($n=1000$).}
The new $n=1000$ set uses generation seed 20261002 and the canonical
$\{3,12\}$ permeability law. Pointwise exact 95\% binomial intervals
for every selector and static coverage are archived with the per-instance
errors. For example, nested seed 42 at $\tau=0.024$ covers 873/1000
examples (87.3\%; interval 85.1--89.3\%), while its static width-16
comparator covers 956/1000 (95.6\%; interval 94.1--96.8\%). These
intervals describe each fitted policy's coverage and are not simultaneous
across the shared test examples. No threshold or budget was retuned using
this set. Two-sample Kolmogorov--Smirnov tests against the $n=624$
calibration pool give $D=0.035$, $p=0.72$ for total variation and
$p=0.29$--$0.999$ for the dense error of the six models; as diagnostics,
these do not establish exchangeability.

Because routing primarily depends on the width-8 error tail, we also
compared its calibration and test distributions. Two-sample KS statistics
for $e_8$ are $D=0.032$--$0.053$ for nested weights ($p=0.21$--$0.54$)
and $D=0.032$--$0.049$ for independent models ($p=0.31$--$0.81$).
These tests do not show a detectable marginal shift, but they have
limited power for tail differences. As a descriptive check, conditional
width-8 failure rates among inputs sent to width 8 were higher on the new
test set in several nested settings; for example, seed 42 at
$\tau=0.024$ had rates 10.6\% on calibration and 14.4\% on test.
This conditional comparison is not itself covered by the marginal
guarantee and does not isolate a cause of the observed coverage gap.

\paragraph{Earlier 200-example test and overlap.}
Before the 1,000-example set, the same locked policies were evaluated on
200 examples generated with seed 20261001. There, nested-selector
coverage was 87.0--90.5\%, with seven of eight feasible cells below
90\%, and every nested pointwise interval included 90\%; independent-selector
coverage was 88.0--95.5\%. Because of a seed-offset collision in the
sharded generator,
examples 1--180 of the 1,000-example set coincide with examples 21--200
of this earlier set. Restricted to the 820 examples that are disjoint
from it, nested-selector coverage is 87.1--90.7\% with six of eight
feasible cells below 90\%; the intervals for the same two cells exclude
90\%; and matched static coverage is higher in all eight. For the
independent family, three of nine cells are below 90\%, against four
on the full set. We report these subsets only as a check and do not pool
results across the two sets.

The resampling and distribution diagnostics below concern the original
canonical test split used during development, not this new test.

\paragraph{Calibration bootstrap.}
We resampled the 624 calibration examples with replacement 2000 times,
recalibrated each time, and evaluated on the fixed test split. The
standard deviation of coverage is 0.7--1.6pp across the examined
combinations. This measures sensitivity to the calibration sample with
the test set held fixed.

\paragraph{Pooled repartitioning.}
We pooled the calibration and test examples and randomly split all 824
into new calibration and test sets 2000 times. At the operating points
that are usually feasible, mean coverage is 90.0--90.2\%, with a
standard deviation of about 2.4pp, and the original partition
reproduces the reported table. About 40--43\% of repartitions fall
below target, which is consistent with sampling variation around 90\%.

These resampling results do not show that the data are exchangeable,
and they cannot undo the use of the test set during development. We do
not combine them across seed--tolerance cells, because the cells share
examples. Seed 43 at $\tau=0.020$ is infeasible in about 84\% of
repartitions, so at that budget calibration with this score often
fails on the pooled data.

\paragraph{Distribution comparisons.}
Two-sample Kolmogorov--Smirnov tests compare the fresh calibration
indices $[400,1024)$ with the test indices $[0,200)$ of the same source
file. For total variation, $D=0.075$ and $p=0.36$; for the dense error,
$p=0.61$, 0.82, and 0.19 for the three seeds. Within pooled-TV
terciles, the width-8 failure rates differ between calibration and test
by $-12.8$ to $+10.0$pp across 27 cells, with no consistent sign. With
roughly 60--74 test examples per tercile, these tests have limited
power, so not finding a difference is only weak evidence of
exchangeability.

\paragraph{Groupwise calibration diagnostic.}
An earlier analysis calibrated the $\eta$-based rule separately within
each TV tercile, following the idea of Mondrian calibration
\citep{vovk2003mondrian}. At $\tau=0.024$, one low-TV group has 97.7\%
coverage with every prediction at width 8, while the high-TV groups are
at or below target and one is infeasible. This analysis used the
checkpoint-selection data and group boundaries chosen from the data, so
it is exploratory.

Here total variation only defines the groups; the rule itself routes on
disagreement. The result therefore says nothing either way about
subgroup behaviour of the binary TV selector, and a formal groupwise
guarantee would need its own calibration design.

\end{document}